\documentclass[]{article}
\usepackage{proceed2e}
\usepackage{graphicx,amsmath,amssymb}
\usepackage[authoryear,round]{natbib}
\usepackage{natbibspacing}
\usepackage{url}
\usepackage{color}

\newcommand{\qed}{\hfill \ensuremath{\Box}}
\newtheorem{theorem}{Theorem}[section]
\newtheorem{lemma}[theorem]{Lemma}

\newtheorem{corollary}[theorem]{Corollary}
\newenvironment{proofsketch}[1][Proof Sketch.]{\begin{trivlist}\item[\hskip \labelsep {\em #1}]}{\end{trivlist}}

\title{Planar Cycle Covering Graphs}
\author{ Julian Yarkony,  Alexander T. Ihler,  Charless C. Fowlkes \\
Department of Computer Science\\ University of California, Irvine\\  \texttt{\{yarkony,ihler,fowlkes\}@ics.uci.edu} }

\begin{document}
\maketitle

\begin{abstract}
We describe a new variational lower-bound on the minimum energy configuration
of a planar binary Markov Random Field (MRF).  Our method is based on adding
auxiliary nodes to every face of a planar embedding of the graph in order to
capture the effect of unary potentials.  A ground state of the resulting
approximation can be computed efficiently by reduction to minimum-weight
perfect matching. We show that optimization of variational parameters achieves
the same lower-bound as dual-decomposition into the set of all cycles of the
original graph.  We demonstrate that our variational optimization converges
quickly and provides high-quality solutions to hard combinatorial problems
10-100x faster than competing algorithms that optimize the same bound.
\end{abstract}

\section{Introduction}
Dual-decomposition methods for optimization have emerged as an extremely powerful tool
for solving combinatorial problems in graphical models.  These techniques can
be thought of as decomposing a complex model into a collection of
easier-to-solve components, providing a variational bound which can then be
optimized over its parameters.  A wide variety of algorithms have been
proposed, often distinguished by the class of models from which subproblems are
constructed, including trees \citep{wainwright05,kolmogorov06}, planar graphs
\citep{Globerson}, outer-planar graphs~\citep{Batra}, k-fans~\citep{Kappes10},
or some more heterogeneous mix of combinatorial
subproblems~\citep[e.g.,][]{torresani08}.

While the class of tree-reweighted methods are now fairly well understood, many
of the same concepts and guidance available for trees are not available for
more general classes of decompositions.  In this paper, we analyze reweighting
methods that seek to decompose binary MRFs into subproblems consisting of
tractable planar subgraphs.  We show that the ultimate building blocks of such
a decomposition are simple cycles of the original graph and that to achieve the
tightest possible bounds, one must choose a set of subproblems that cover all
such cycles. Cycles in planar-reweighted decomposition thus play a role
analogous to trees in tree-reweighted decompositions.

There are various techniques for enforcing consistency over cycles in an MRF.
For example, one can triangulate the graph and introduce constraints over all
triplets in the resulting triangulation.  However, this involves $O(n^3)$ 
constraints which is impractical in large-scale inference problems.  A more
efficient route is to only add a small number of constraints as needed, e.g.,
using a cutting-plane approach~\citep{Sontag07}.  

The contribution of this
paper is a graphical construction for a new variational bound that enforces the
constraints over {\em all cycles} in a planar binary MRF with only a constant
factor overhead.  This representation is very simple and efficient to optimize,
which we demonstrate in experimental comparisons to existing state-of-the-art,
cycle-enforcing methods where we achieve substantial performance gains.

%For example, many of these subproblem types lack guidance as to what subproblems to
%include, or how to 
%We focus on a particular class of problems consisting of binary-valued random variables
%whose distribution is specified by a planar, pairwise graphical model

%While the class of tree-reweighted methods are becoming fairly well understood, many of the same
%concepts and guidance available in the tree case are not available for more general classes
%of decompositions.  In this work, we examine the classes of reweighted methods using planar
%and outer-planar components.  Both have been proposed in the literature, but without any
%clear guidance of what problems to select or how to guarantee an optimally tight decomposition bound.
%We show that for the case of outer-planar decompositions, there exists a sufficient and necessary
%set of problems which must be included for optimality, namely the set of all single cycles in the model.
%For the class of binary, planar graphical models with external fields, we also propose a new
%\emph{single-problem} relaxation that we show fully contains this set and is thus sufficient
%to attain the bound provided by any decomposition into outer-planar subproblems.
%We show in experiments that our single-problem method converges rapidly and provides 
%high-quality solutions to difficult combinatorial optimization problems.
\begin{figure*} [t]
\center
\includegraphics[width=0.6\linewidth]{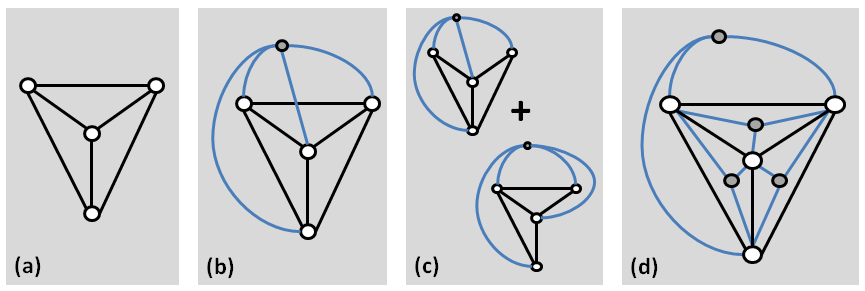}
\caption{(a) shows a standard planar MRF which is represented by an energy function
containing unary and pairwise potentials (b) shows an equivalent MRF in which the 
unary terms have been replaced by an auxiliary node (square). Both (a) and (b) are intractable
in general.  (c) shows a decomposition which gives a lower-bound on the ground-state
of (a) by using a collection of outer-planar graphs whose ground states can be
computed efficiently using minimum-weight perfect matching. (d) shows the new
lower-bound construction introduced in this paper which uses multiple auxiliary
nodes, one for each face of the original graph.}
\label{fig:overview}
\end{figure*}

\section{Exact Inference for Binary Outer-planar MRFs}

Consider the energy function $E(X)$ associated with a general binary MRF
defined over a collection of variables $(X_1,X_2,\ldots) \in \{0,1\}^N$ with
specified unary and pairwise potentials.  It is straightforward to show that
any such MRF can be reparametrized up to a constant using pairwise disagreement
costs $\theta_{ij}$ along with unary parameters $\theta_{i}$ %(see e.g.,
\citep[see, e.g.,][]{kolmogorov04,SchraudolphTR}.  The energy function can thus be written as
\begin{equation}
E(X,\theta) = \sum_{i>j} \theta_{ij} [X_i\not=X_j] + \sum_{i} \theta_i [X_i\not=0] 
\label{eqn:energyfun}
\end{equation}
where $[\cdot]$ is the indicator function and we have dropped any constant
terms.\footnote{We assume in the rest of this paper that all MRFs are parameterized 
in this manner.  In particular an MRF without unary parameters is one in which
all the pairwise terms are symmetric.}

We can express such an energy function without including any unary terms
by introducing an auxiliary variable $X_0$ and replacing the unary
terms with pairwise connections to $X_0$ so that
\begin{align}
E_1(X,\theta) = \sum_{i>j} \theta_{ij} [X_i\not=X_j] + \sum_{i} \theta_i [X_i\not=X_0]
\label{eqn:ising}
\end{align}
If we fix $X_0=0$, then $E_1$ is clearly equivalent to our original energy
function $E$.  Since the potentials in $E_1$ are symmetric, for any state $X =
(X_0,X_1,\ldots)$, there is a state $\bar X$ with identical energy, given by
flipping the states of every $X_i$ including $X_0$. Thus any $X$ that minimizes
$E_1$ can be easily mapped to a minimizer of $E$.

Minimizing the energy function $E_1$ can be interpreted as the problem of
finding a bi-partition of a graph ${\mathcal G}_1$ which has a vertex $i$
corresponding to each variable $X_i$ and edges for any pair $(i,j)$ with
$\theta_{ij}\not=0$.  The cost of a partition is simply the sum of the weights
$\theta_{ij}$ of edges cut.  Given a minimal weight partition, we can find a
corresponding optimal state $X$ by assigning all the nodes in the partition
containing $X_0$ to state $0$ and the complement to state $1$.  Since the edge
weights $\theta_{ij}$ may be negative, such a minimal weight cut is typically
non-empty.

While minimizing $E(X,\theta)$ is computationally intractable in
general~\citep{Barahona82}, a clever construction due to
%Kasteleyn~
\citet{Kasteleyn1,Kasteleyn2} and 
%Fisher~
\citet{Fisher1,Fisher2} allows
one to find minimizing states when the graph corresponding to $E_1$ is planar.
This is based on the complementary relation between states of the nodes $X$ and
perfect matchings in the so-called expanded dual of the graph ${\mathcal G}_1$.  
A minimizing state for a planar problem can thus be found efficiently, e.g. using
Edmonds' {\it blossom} algorithm~\citep{Edmonds1965} to compute minimum-weight
perfect matchings.\footnote{Matchings in planar graphs can be found somewhat
more efficiently than for general graphs which yields the
best known worst-case running time of $O(N^{3/2} \log N)$ for
max-cut in planar graphs~\citep{ShihWuKuo90}.}
We use the Blossom V implementation of \citet{Kolmogorov09} which is
quite efficient in practice, easily handling problems with a million nodes in a
few seconds.  Furthermore, for planar problems, one can also compute the
partition function associated with $E$ in polynomial time.  See the report of
\citet{SchraudolphTR} for an in-depth discussion and
implementation details.

While this reduction to perfect matching provides a unique tool for energy
minimization and probabilistic inference, the requirement that ${\mathcal G}_1$ be planar
is a serious restriction.  In particular, even if the original graph ${\mathcal G}$ 
corresponding to $E$ is planar, e.g., in the case of the grid graphs commonly
used in computer vision applications, ${\mathcal G}_1$ is typically not, since the
addition of edges from every node to the auxiliary node $X_0$ renders the graph
non-planar.  Assuming arbitrary values of $\theta_i$, 
those energy functions $E$ to which this method can be applied are
exactly the set whose graphs ${\mathcal G}$ are {\it outer-planar}. An outer-planar
graph is a graph with a planar embedding where all vertices share a common face
(e.g., the exterior face). For such a graph, every vertex can be connected to a
single auxiliary node placed inside the common face without any edges crossing
so that the resulting graph ${\mathcal G}_1$ is still planar. 
See examples in Figure \ref{fig:overview}.%
\footnote{Note that outer-planar graphs have treewidth two and hence the
minimum energy solution can also be found efficiently using the
standard junction tree algorithm. However, the reduction to matching is still
of interest for general planar graphs without unary potentials, which have a
treewidth of $O(\sqrt{N})$.}

\section{Inference with Dual Decomposition}

Dual decomposition is a general approach for leveraging such
islands of tractability in order to perform inference in more general MRFs. The
application of dual decomposition to inference in graphical models was
popularized by the work of \citet{wainwright03,wainwright05}
on Tree-Reweighted Belief Propagation (TRW).  TRW finds an optimal
decomposition of an MRF into a collection of tree-structured problems where
exact inference is tractable.  More formally, let $t$ index a collection of
subproblems defined over the same set of variables $X$ and whose parameters sum
up to the original parameter values, so that $\theta = \sum_t \theta^t$.  The
energy function is linear in $\theta$ so we have
\begin{align}
E_{MAP} &= \min_X E(X,\Theta) = \min_X \sum_t E(X,\Theta^t) \label{eqn:dd1} \\  
&\qquad\geq \max_{\sum_t \theta^t=\theta} \sum_t \min_{X^t} E(X^t,\Theta^t)
\label{eqn:dd}
\end{align}
The inequality arises because each subproblem $t$ is solved independently and
thus may yield different solutions.  On the other hand, if the solutions to
the sub-problems all happen to agree then the bound is tight.  The %variational
problem of maximizing the lower-bound over possible decompositions
$\{\theta^t\}$ is convex and when inference for each sub-problem is tractable
(for example, $\theta^t$ is tree-structured) the bound can be optimized efficiently
using message passing (fixed-point iterations) based on computing 
min-marginals in each subproblem~\citep{wainwright03} or by projected
subgradient methods~\citep{komodakis07}.

A powerful tool for understanding the minimization in Equation \ref{eqn:dd} is
to work with the Lagrangian dual.  Equation~\ref{eqn:dd1} is an integer linear
program over $X$, but the integrality constraints can be relaxed to a linear
program over continuous parameters $\mu$ representing min-marginals which are
constrained to lie within the \emph{marginal polytope}, $\mu \in {\mathbb
M}({\mathcal G})$.  The set of constraints that define ${\mathbb M}({\mathcal G})$ are
a function of the graph structure $\mathcal G$ and are defined by an (exponentially
large) set of linear constraints that restrict $\mu$ to the set of
min-marginals achievable by some consistent joint distribution
\citep[see][]{wainwright08}.  Lower-bounds of the form in Equation \ref{eqn:dd}
correspond to relaxing this set of constraints to the intersection of the
constraints enforced by the structure of each subproblem.  For the
tree-structured subproblems of TRW, this relaxation results in the so-called
\emph{local} polytope ${\mathbb L}({\mathcal G})$ which enforces marginalization
constraints on each edge.  Since ${\mathbb L}({\mathcal G})$ is an outer bound on
${\mathbb M}({\mathcal G})$, minimization yields a lower-bound on the original
problem.  For any relaxed set of constraints, the values of $\mu$ may not
correspond to the min-marginals of any valid distribution, and so are referred
to as pseudo-marginals.

One can tighten the bound in Equation~\ref{eqn:dd} by adding additional
subproblems to the primal (or equivalently constraints to the dual) which
enforce consistency over larger sets of variables.  This has been explored,
e.g. by \citet{Sontag07} who suggest adding cycle
inequalities to the dual which enforce consistency of pseudo-marginals around a
cycle. Since there are a large number of potential cycles present in the graph,
Sontag suggests either using a cutting plane algorithm to successively add
violated cycle constraints~\citep{Sontag07} or to only add small cycles such as 
triplets or quadruplets~\citep{Sontag08} that can be enumerated with relative ease
and optimized using local message passing rather than general LP solvers.

For binary problems, it is natural to consider replacing Wainwright's tree
subproblems with tractable outer-planar subgraphs.  This has been explored by
\citet{Globerson} and \citet{Batra} who
proposed decomposing a graph into a set of planar graphs for the
purposes of estimating the partition 
function\footnote{More precisely, \citet{Globerson} consider the inclusion
of any binary, planar subgraph of ${\mathcal G}_1$. This may include subgraphs with
treewidth greater than two.}
and minimum energy state
respectively.  
For energy minimization, it is well-known that any set of subproblems
that cover every edge is sufficient to achieve the TRW bound; but what is the
best set of planar graphs to use?   Is it necessary to use all outer-planar or even all
planar subgraphs?
It turns out that %, for the purposes of finding a minimum energy state,
the set of all outer-planar or planar subgraphs is equivalent to the set of
all cycle constraints in ${\mathcal G}$, which can be enforced by any so-called
\emph{cycle basis} of the graph.  This observation leads to algorithms such
as reweighted perfect matching \citep{RPM}, which explicitly constructs
a set of subproblems that form a complete cycle basis, or incremental algorithms to
enforce cycle constraints~\citep{Sontag07,Sontag08,Komodakis08}.

%While decomposing $E$ into planar graphs appears to work quite
%well experimentally there has been relatively little analysis.  The nicely
%developed notions of weak tree agreement, optimality conditions and duality
%that have been worked out for TRW~\citep{kolmogorov06} are missing.  In
%particular, unlike TRW, it is not clear what collection of outer-planar graphs
%to use as subproblems.  Both~\citet{Globerson} and \citet{Batra} note that the
%decomposition must cover every edge in the original graph at least once, but
%this is clearly insufficient for achieving the tightest bound possible.  For
%example, the set of spanning trees is a planar decomposition that covers every
%edge, but only yields the bound of TRW, with none of the benefits demonstrated
%by decomposition into graphs that contain cycles. What is the best collection
%of planar graphs to use?  Is it necessary to use all outer-planar or even all
%planar subgraphs?

In the following sections, we focus on the case in which the original MRF is
planar but the addition of the auxiliary unary node makes it non-planar.  We
describe a novel, compactly expressed variational approximation.  We then prove
that it achieves as tight a bound as decomposition into any collection of cycles or
outer-planar graphs.  This also gives a relatively simple proof that the tightest bounds
achievable by sets of planar, outer-planar, or cycle subproblems are equivalent,
and that the set of subproblems that are necessary and sufficient to achieve this
bound form a cycle basis, i.e., cover every chordless cycle in the original graph 
at least once.

%of the set of
%tractable outer-planar subproblems that are necessary and sufficient to achieve
%this bound, namely that it must form a cycle basis, i.e.,
%%any outer-planar decomposition must 
%cover every chordless cycle in the original
%graph at least once.
%% in order to achieve the tightest possible bound for a
%%planar MRF.
%n immediate characterization of the set
%of tractable outer-planar subproblems that are necessary and sufficient, namely
%any outer-planar decomposition must cover every chordless cycle in the original
%graph at least once in order to achieve the tightest possible bound for a
%planar MRF.

\section{Planar Cycle Coverings}

\begin{figure*} [t]
\center
\includegraphics[width=0.6\linewidth]{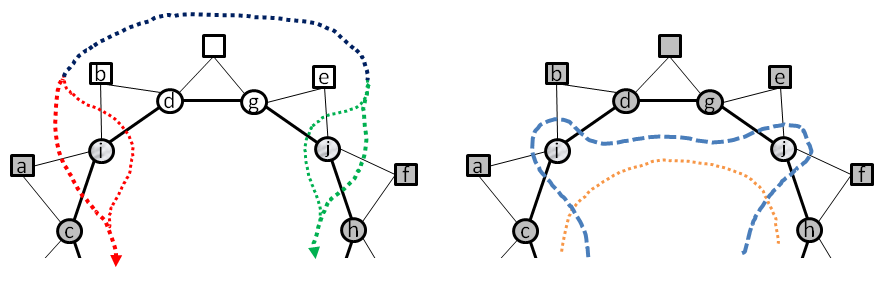}
\caption{Demonstration that the minimal energy of a cycle is equal to the
maximum lower-bound given by an approximation in which unary potentials are
represented by a decoupled set of auxiliary variables (squares).  At optimality of the
variational parameters, all six cuts depicted must have equal energies and
thus it is possible to choose a ground-state in which all the duplicate copies
of the auxiliary node are in the same state.}
\label{fig:cuts}
\end{figure*}

Consider a planar embedding of the graph ${\mathcal G}$ corresponding to an MRF.
Since we cannot directly connect the unary node $X_0$ to every node in the graph
without losing planarity, we propose the following relaxation.  For each face
$f$ of ${\mathcal G}$ add an independent copy of the unary node $X_0^f$ and connect
it to all vertices on the boundary of the face with weights $\theta^f_i$.
Let $N_i$ be the set of unary node copies attached to node $i$.
We split the original unary potential $\theta_i$ across all the unary 
face nodes connected to $i$ while maintaining the constraint that $\sum_{f \in N_i}
\theta^f_i = \theta_i$; see Figure \ref{fig:overview}(d).  Using this system
we have the following relaxation
\begin{align}
%\min_X E(X,\Theta) 
E_{MAP} &= \min_{X:X_0^f = X_0} \sum_{i>j} \theta_{ij} [X_i\not=X_j] + \sum_{i,f} \theta^f_i [X_i\not=X^f_0] \nonumber\\
&\geq \min_X \sum_{i>j} \theta_{ij} [X_i\not=X_j] + \sum_{i,f} \theta^f_i [X_i\not=X^f_0]
\label{eqn:relax}
\end{align}
The inequality arises because we have dropped the constraint that all copies of
$X_0$ take on the same value.  On the other hand, since the graph corresponding
to the relaxation in Equation~\ref{eqn:relax} is planar, we can compute the
minimum exactly.  Furthermore, we have freedom to adjust the $\theta^f_i$
parameters so long as they sum up to our original parameters.  This yields the
variational problem
\begin{equation}
E_{PCC} = \!\!\max_{\theta : \sum_f \theta^f_i = \theta_i}\min_X \sum_{i>j} \theta_{ij} [X_i\!\ne\!X_j] + \sum_{i,f} \theta^f_i [X_i\!\not=\!X^f_0]
\end{equation}
where $E_{MAP} \geq E_{PCC}$.  We refer to this construction as a {\em planar
cycle covering} of the original graph since the singular potentials for each
face cycle are covered by some auxiliary node (and as we shall see, all other
cycles also are covered in a precise sense).  Although this planar
decomposition includes duplicate copies of nodes from the original problem, it
differs in that there are not multiple independent subproblems but just a
single, larger planar problem to be solved.  This is in some ways analogous to
the work of \citet{yarkony10} which replaces the collection of spanning trees in
TRW with a single ``covering tree''.

%\subsection{Subgradient optimization}

As with dual decomposition, the parameters may be optimized using subgradient
or marginal fixed-point updates.  For example, the subgradient updates for
$\theta^f_i$ at a given setting of $X$ can be easily computed by taking a
gradient and enforcing the summation constraint.  This yields the update rule
\begin{equation}
\theta^f_i = \theta^f_i + \lambda \left([X_i\not=X^f_0] - \frac{1}{|N_i|} \sum_{g \in N_i} [X_i\not=X^g_0]\right)
\label{eqn:subgrad}
\end{equation}
where $|N_i|$ is the number of auxiliary face nodes attached to $X_i$ and $\lambda$ is a
stepsize parameter.  After each such gradient step, one must recompute the 
optimal setting of $X$ which can be done efficiently using perfect matching.

The subgradient update lends itself to a simple interpretation.  If $X^f_0$
disagrees with $X_i$ but the other neighboring copies $\{X^g_0\}$ do not, then the
cost for $X^f_0$ and $X_i$ disagreeing is increased. On the other hand, if all
the copies $\{X^g_0\}$ take on the same state then the update leaves the
parameters unchanged. 

\section{Cycle Decompositions and Cycle Covering Bounds}

In this section, we show that the planar cycle cover bound $E_{PCC}$ for any
planar binary MRF ${\mathcal G}$ is equivalent to the lower-bound given by
decomposition into the collection of all cycles of ${\mathcal G}$. 

For a given planar binary MRF with graph ${\mathcal G}$, consider the bound
$E_{CYCLE}$ given by decomposing the MRF into the collection of all cycles of
${\mathcal G}$.  By optimizing the allocation of parameters across these
subproblems one produces a lower-bound that is generally tighter than that
given by TRW and related algorithms since the subproblems can correctly account for
the energy of frustrated cycles that is approximated in the tree-based bound.
In fact, for planar graphs without unary potentials adding cycle subproblems is
enough to make the lower-bound tight.

\begin{lemma}
\label{lemma:cycle}
The lower-bound $E_{CYCLE}$ given by the optimal cycle decomposition of a
planar MRF with no unary potentials is tight.
\end{lemma}

For such an MRF the set of states corresponds exactly with the set of edge
incidence vectors representing cuts in the graph. The convex hull of this 
set is known as the cut polytope.  The connection between the cut polytope and
the cycle decomposition is seen by taking the Lagrangian dual of the
lower-bound optimization which yields a constrained optimization of the edge
incidence vectors (pseudo-marginals) over a polytope defined by cycle
inequalities.  For planar graphs (or more generally graphs containing no $K_5$
minor), the set of cycle inequalities is sufficient to completely describe
the cut polytope.  See \citet{BarahonaMahjoub86} for proof and related
discussion by~\citet{Sontag07}. Just as local edge consistency implies
global consistency for a tree, cycle consistency implies global consistency
for a planar binary MRF without unary potentials.

While the number of simple cycles grows exponentially in the size of the graph
for general planar graphs, it is still possible to solve such a problem in
polynomial time.  It is not in fact necessary to include every cycle subproblem
but simply a subset which form a cycle basis~\citep{Barahona93}.  Furthermore,
there exists an efficiently computable witness for identifying a violated
cycle~\citep{BarahonaMahjoub86}.  \citet{Sontag07} use this as the basis for a
cutting plane method which successively adds cycle constraints to the dual.%
\footnote{It is important to note that a cycle basis for ${\mathcal G}_1$
is not sufficient to achieve the bound $E_{CYCLE}$ given by the collection of
all cycles in ${\mathcal G}$ since a cycle in ${\mathcal G}$ corresponds to 
a wheel in ${\mathcal G}_1$.}

We would now like to consider cycles in MRFs which do have unary potentials.
We start with the simplest case of a single cycle.
\begin{lemma}
The minimum energy of a single cycle is the same as the maximum lower-bound
given by the graph in which the unary potentials have been replaced by a
collection of auxiliary nodes (one for each edge in the cycle) where each node
in the cycle is connected to the pair of auxiliary nodes corresponding to its
incident edges.
\label{lemma:split}
\end{lemma}
\begin{proofsketch}

Figure~\ref{fig:cuts} provides a visualization of the set of auxiliary
nodes (squares) added to the cycle (circles).  We refer to this as the ``saw'' graph.
Suppose we have optimized the decomposition of unary parameters across the
auxiliary node connections to maximize the lower-bound.  We claim that at the
optimal decomposition, there always exists a minimal energy configuration such
that all the auxiliary nodes take on state $0$, making the bound equivalent
to the cycle with a single auxiliary node.

Suppose we choose a minimum energy configuration of the graph but the duplicate
auxiliary nodes take on mixed states.  Start at some point along the cycle
where there is an auxiliary node in state 0 and proceed clockwise until we find
an auxiliary node in state 1.  As we continue around the cycle we will
encounter some later point at which the auxiliary nodes return to being in
state 0. This is most easily visualized in terms of the cut separating $0$ and
$1$ nodes as shown in Figure \ref{fig:cuts}. 

Let $X_i$ be the first node
which is attached to a pair of disagreeing auxiliary nodes $X^a_0,X^b_0$ and
$X_j$ be the second attached to $X^e_0,X^f_0$. 
Consider the four possible cuts highlighted in red and green in Figure
\ref{fig:cuts}.  At the optimal decomposition of the parameters, it must be the case
that these paths have equal costs.  If not, then we could transfer
weight (e.g. from $\theta^a_i$ to $\theta^b_i$) and increase the energy, contradicting
optimality. Let $C_1 = (\theta_{ic} + \theta^a_i) = (\theta_{id} + \theta^b_i)$
and $C_2 = (\theta_{jh} + \theta^f_j) = (\theta_{jg}+\theta^e_j)$.  If one
of the four cuts shown is minimal then it must be that $C_1 + C_2 \leq 0$, 
otherwise the path which cuts none of these edges (orange) would be preferred.
However, if $C_1+C_2<0$ then there is yet another cut (blue) which
would achieve an energy that is lower by a non-zero amount $(C_1+C_2)$ by
cutting both sets of edges.  Therefore, it must be the case that $C_1+C_2 = 0$
and thus either orange or blue cuts also represents a minimal configuration
that leaves the collection of auxiliary nodes in state $0$.  A similar line of
argument works for the cases when $X_c = 1$ or $X_h = 1$ or both.

We are thus free to flip the states of the block of disagreeable auxiliary
nodes and their neighbors on the cycle without changing the energy.  We can
then continue around the cycle in this manner until all copies of the auxiliary
nodes are in state $0$ as desired.
\qed

%The given configuration has a cost given by one of 4
%possibilities $\theta^a_0 + \theta^c_0 + C$, $\theta^b_0 + \theta^c_0 + C$,
%$\theta^a_0 + \theta^d_0 + C$, $\theta^b_0 + \theta^d_0 + C$ where $C$ is a
%constant that capture the cost of the remainder of the cut.
\end{proofsketch}

We are now ready to give the main result of this section.

\begin{figure*} [t]
\center
\includegraphics[width=0.8\linewidth]{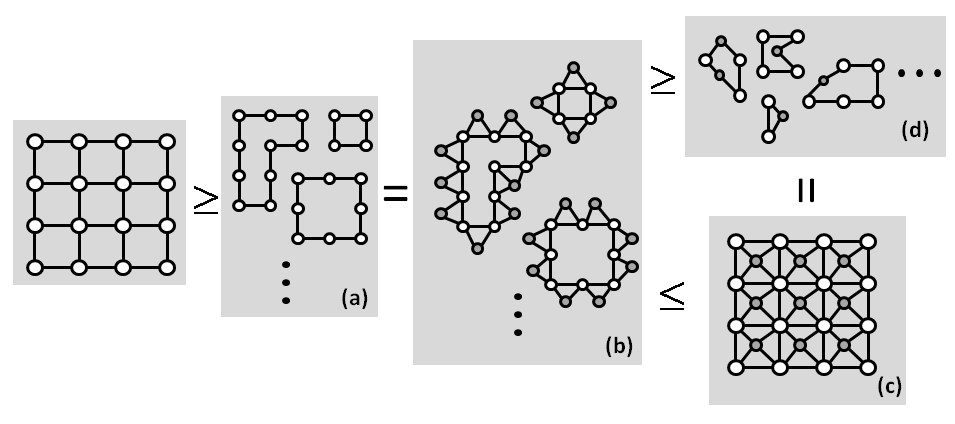}
\caption{Graphical depiction of Theorem~\ref{thm:equiv} demonstrating that 
the planar cycle covering graph enforces constraints over all cycles of 
the original graph. (a) depicts the lower bound $E_{CYCLE}$ based on a
decomposition into the collection of all simple cycles of the original graph.
Lemma~\ref{lemma:split} shows that this bound is equivalent to the bound given
by a corresponding collection of graphs (b) in which unary potentials are 
captured by multiple auxiliary nodes placed along each edge.  Since
every one of these graphs is a subgraph of the planar cycle covering graph (c)
their minimum energy must be less than $E_{PCC}$.  Finally, since the planar
cycle covering graph (c) has no unary potentials, it is equal to its collection
of cycles which are themselves all subgraphs of (b).}
\label{fig:proof}
\end{figure*}

\begin{theorem}
\label{thm:equiv}
The lower-bound given by the planar cycle covering graph is equal
to the lower-bound given by decomposition into the collection of 
all cycles so that $E_{PCC} = E_{CYCLE}$.
\end{theorem}
\begin{proofsketch}
We proceed by showing a circular sequence of inequalities.
Figure~\ref{fig:proof} provides a graphical overview.  Take the set of cycles
which yield the bound $E_{CYCLE}$.  We can apply Lemma~\ref{lemma:split} to
transform each cycle subproblem into a corresponding ``saw'' containing an
auxiliary node for each edge while maintaining the bound.  We then observe
that every such augmented cycle is a subgraph of the planar cycle covering
graph.  As with any such decomposition into subgraphs, the minimal energy of
the cycle covering graph must be at least as large as the sum of the minimal
subgraph energies and hence $E_{CYCLE} \leq E_{PCC}$.
%(Equivalently, any graph enforces all consistency constraints enforced by its
%subgraphs, and thus... 
%since the set of augmented cycles
%are exactly a subproblem decomposition of the PCC graph (individual cycles
%might take on different states in the minimization but they are forced to agree
%in the PCC graph). 
On the other hand, since the PCC graph is now a planar binary
MRF with no unary terms, by Lemma~\ref{lemma:cycle} we can decompose it 
exactly into the collection of its constituent cycles with no loss in the
bound.  Finally each of these cycles is itself a subgraph of some augmented
cycle and hence we must also have that $E_{CYCLE} \geq E_{PCC}$, proving
equality. \qed
\end{proofsketch}

\citet{Batra} and \citet{Globerson} both propose
decomposing a binary MRF into a set of tractable planar graphs. Based on the
previous result, we can clearly see that the best achievable bound under
such a decomposition must include a subproblem that covers every
chordless cycle in the original graph. If consistency along a particular cycle is
not enforced we can always arrange parameters so that the resulting bound
is arbitrarily bad.  We also show the converse, that outer-planar decomposition
can do no better than the set of cycles.

\begin{corollary}
The best lower-bound achieved by any outer-planar decomposition for a planar
MRF is no larger than $E_{PCC}$.
\end{corollary}
\begin{proofsketch}
Take any outer-planar decomposition of a planar MRF.  We first note that an
outer-planar graph may be decomposed into a forest of blocks consisting of
either biconnected components or individual edges, where blocks are connected
by single vertices (cut vertices).  Each biconnected component in turn has a
dual graph which is a tree, meaning it consists of face cycles which have one
edge in common (see e.g., \citet{Syslo} for a more in-depth discussion).

We first split apart the forest into blocks.  Consider any pair of blocks
connected at a single cut vertex $X_i$.  To split them, we introduce copies
$X^1_i$ $X^2_i$ of the cut vertex which are allowed to take on independent states. The unary
parameter $\theta_{i}$ is shared between these two copies with the constraint
that $\theta^1_{i} + \theta^2_{i} = \theta_{i}$. There exists an optimal
decomposition of $\theta_{i}$ which assures the two nodes share an optimizing
configuration.  For, suppose to the contrary that the optimal decomposition
yielded a minimum energy configuration where $X^1_i$ and $X^2_i$ took on
different states, say $X^1_i=0$ and $X^2_i=1$. Then, shifting weight from
$\theta^1_{i}$ to $\theta^2_{i}$ would drive up the energy of such a
disagreeing configuration, contradicting optimality of the decomposition.

Once blocks have been split apart, we may apply essentially the same argument to
split each biconnected component into its constituent face cycles.  Consider
the pair of neighboring nodes $X_i$,$X_j$ which are split into
$X^1_i$,$X^2_i$,$X^1_j$, and $X^2_j$.  At the optimal decomposition of the
parameters $\theta_i, \theta_j, \theta_{ij}$, it again must be the case that
the copies of the duplicated edge must share at least one optimizing configuration.  If
not then the parameters could be redistributed by removing weight from one or
more unused states in one copy and adding it to the set of optimizing states
for the other copy. This would increase the energy and thus contradict
optimality of the decomposition.

Thus any outer-planar decomposition is equivalent to a bound given by the set
of constituent cycles and edges.  Every one of these subproblems is a subgraph
of the cycle covering graph and so the bound can be no tighter than the PCC
graph bound.\qed
\end{proofsketch}

\section{Experimental Results}
We demonstrate the performance of the planar cycle cover bound on randomly generated
Ising grid problems, and compare against two state-of-the-art approaches:
max-product linear programming (MPLP) with incrementally added cycles \citep{Sontag08} 
and reweighted perfect matching (RPM) \citep{RPM}.

Each problem consists a grids of size $N$x$N$  with pairwise potentials drawn from a
uniform distribution $\theta_{ij}\sim U(-1,1)$.  The unary potentials are generated
from a uniform distribution $\theta_i \sim U(-a,a)$, where the magnitude $a$ 
determines the difficulty of the problem.  Large values are relatively
easy to solve, since each variable has strong local information about its optimal
value; as $a$ becomes smaller the problems typically become more difficult.
We generate three categories of problem, ``easy'' ($a=3.2$), ``medium'' ($a=0.8$),
and ``hard'' ($a=0.2$), and show the results on each class of problem separately.
To make it easy to test convergence, we scaled the weights by $500$ and rounded
them to integers. Thus a gap of less than $1$ between lower and upper bounds 
provides a certificate of optimality.

We implemented the PCC bound using the Blossom V implementation of \citet{kolmogorov04}.
At each step $t$ we obtain both a lower-bound $E_{PCC}^t$ and a configuration of
$X=[X_1,\ldots,X_N]$ and the copies $\{X_0^f\}$.  We compute the energy of two
possible joint solutions, $X$ and its complement $\bar X$, and save the best
solution found so far and its energy $\hat E^t$ as a current upper bound.
The variational parameters are updated
using the projected sub-gradient given in Equation~\ref{eqn:subgrad}, and the
step size $\lambda$ is chosen using Polyak's step size rule,
i.e., given sub-gradient $g(\theta)$ we choose $\lambda=\frac{1}{2}(\hat E^t - E_{PCC}^t)/\|g\|^2$.
The incremental update feature of Blossom V is used to speed up successive 
optimizations as the variational parameters are modified.

For both MPLP and RPM, we used the original authors' code available online. 
MPLP first runs an optimization corresponding to the tree-reweighted lower bound (TRW),
then successively tightens this bound by trying to identify cycles whose constraints
are significantly violated and adding those subproblems to the collection.  For grids,
it enumerates and checks each square of four variables; we modified the code slightly to
ensure that any given square is added only once.  Because weak tree agreement can lead
to suboptimal fixed points in MPLP, we tried both the standard message updates and a
version which used subgradient steps, but found little difference and report only
the fixed point update results.  We also note that because this implementation of 
MPLP explicitly enumerates only a subset of cycles, the MPLP implementation may
not provide the tightest possible lower-bound, an effect we observe in our
experiments.

For RPM, we used the author's implementation \emph{IsInf}, which uses a
bundle-trust optimization subroutine for its subgradient updates.  IsInf does
not compute upper bounds (proposed solutions) frequently; in plots showing the
change in bounds over time we modified the code to also return such a solution,
but used the default behavior for our timing comparisons.

\begin{figure*}[t]
\center
\includegraphics[width=2.2in]{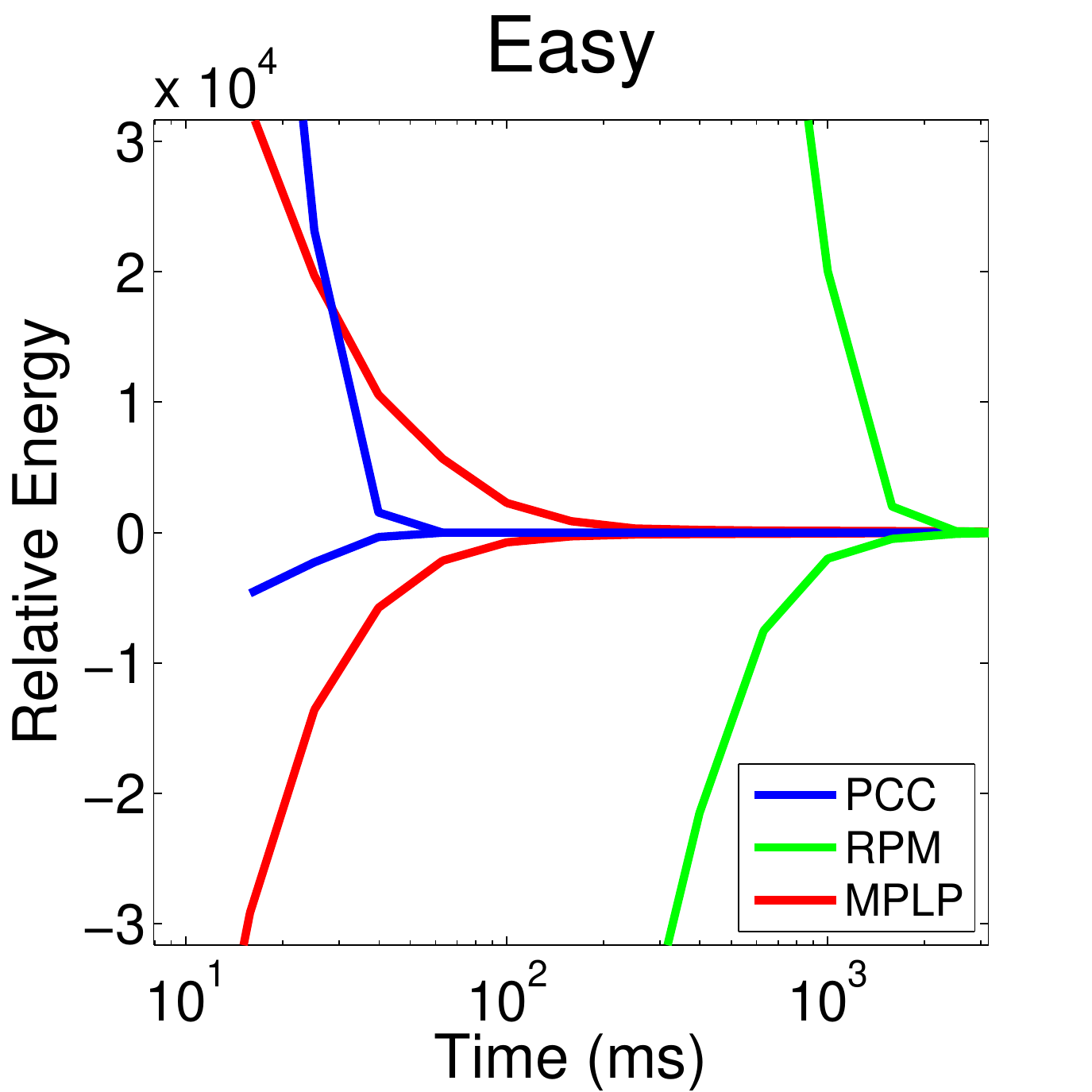}
\includegraphics[width=2.2in]{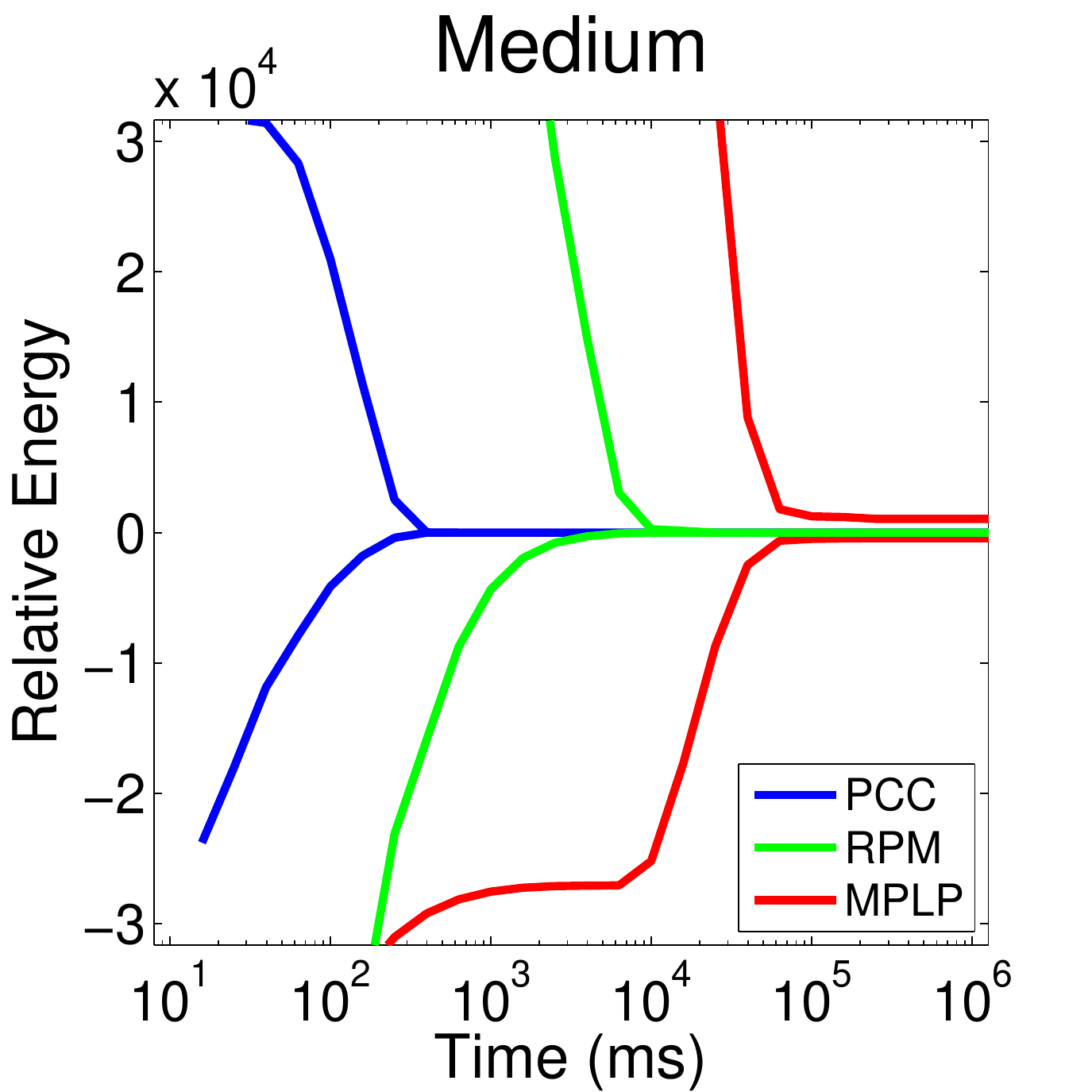}
\includegraphics[width=2.2in]{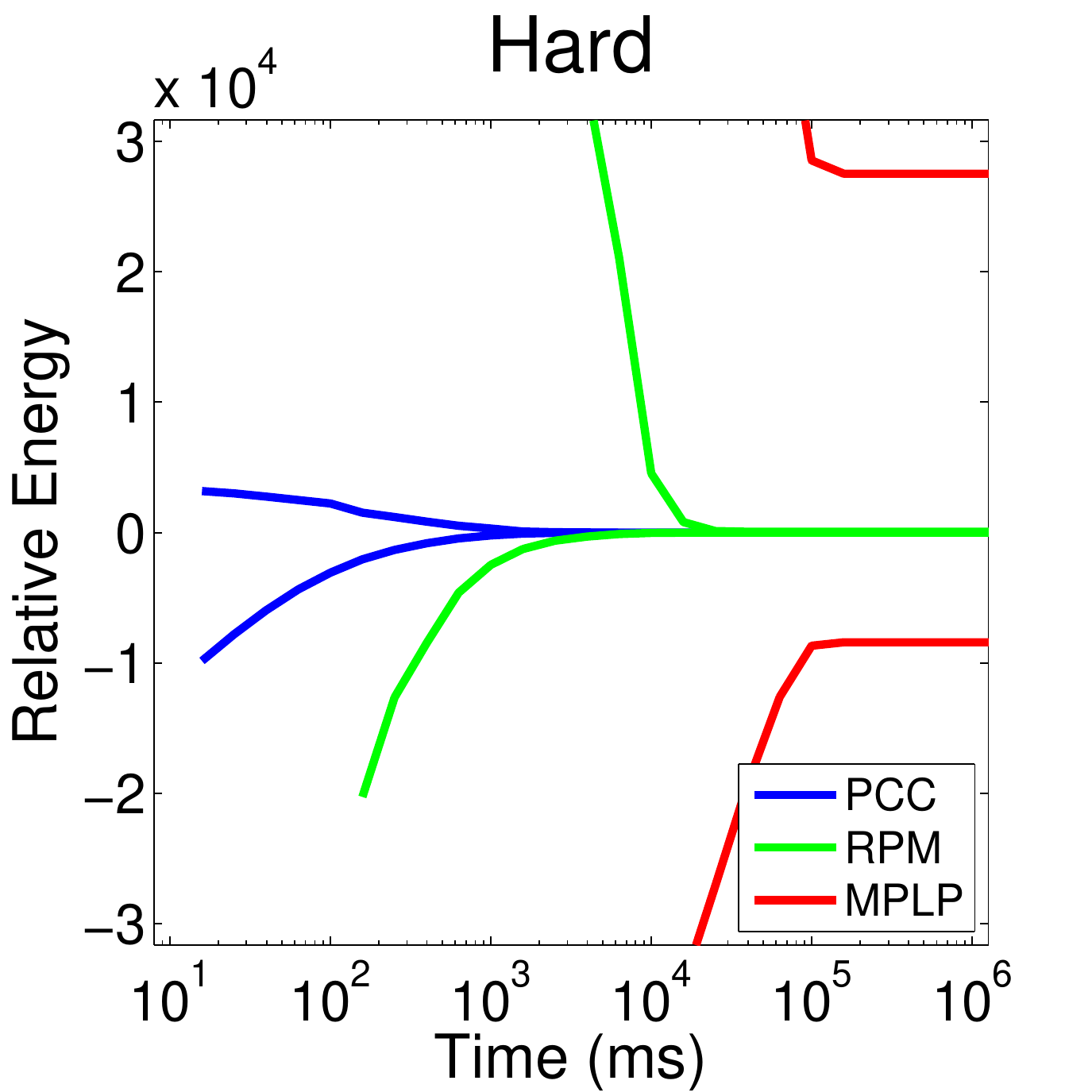}
\caption{Average convergence behavior of lower- and upper-bounds for randomly
generated 32x32 Ising grid problems.  We compare PCC, the planar cycle cover
bound (blue) to RPM (green) and  MPLP (red) for easy, medium and hard problems.
The problem difficulty is controlled by the relative influence of unary and
pairwise potentials. Energies are averaged over 10 random problem instances and
plotted relative to a MAP energy of $0$.}
\label{fig:results}
\end{figure*}

Figure~\ref{fig:results} shows the upper and lower bounds found by each
algorithm as a function of time, for a single $32\times 32$ problem instance
from each of the three categories.  For the ``easy'' problem, all three methods
find and verify the optimal solution (zero duality gap); in this case, MPLP
converges more quickly than RPM, and PCC is faster still.  For the ``medium''
problem, we see that MPLP converges more slowly and to a small duality gap,
with RPM slightly faster and PCC still fastest.  For the ``hard'' problem,
MPLP has a large duality gap; in this case RPM and PCC still converge to and
verify the optimum.  In all cases, PCC is significantly faster than the other
methods.

\begin{figure*}[t]
\center
\begin{tabular}{ccc}
\hspace{-1em}\includegraphics[width=2.2in]{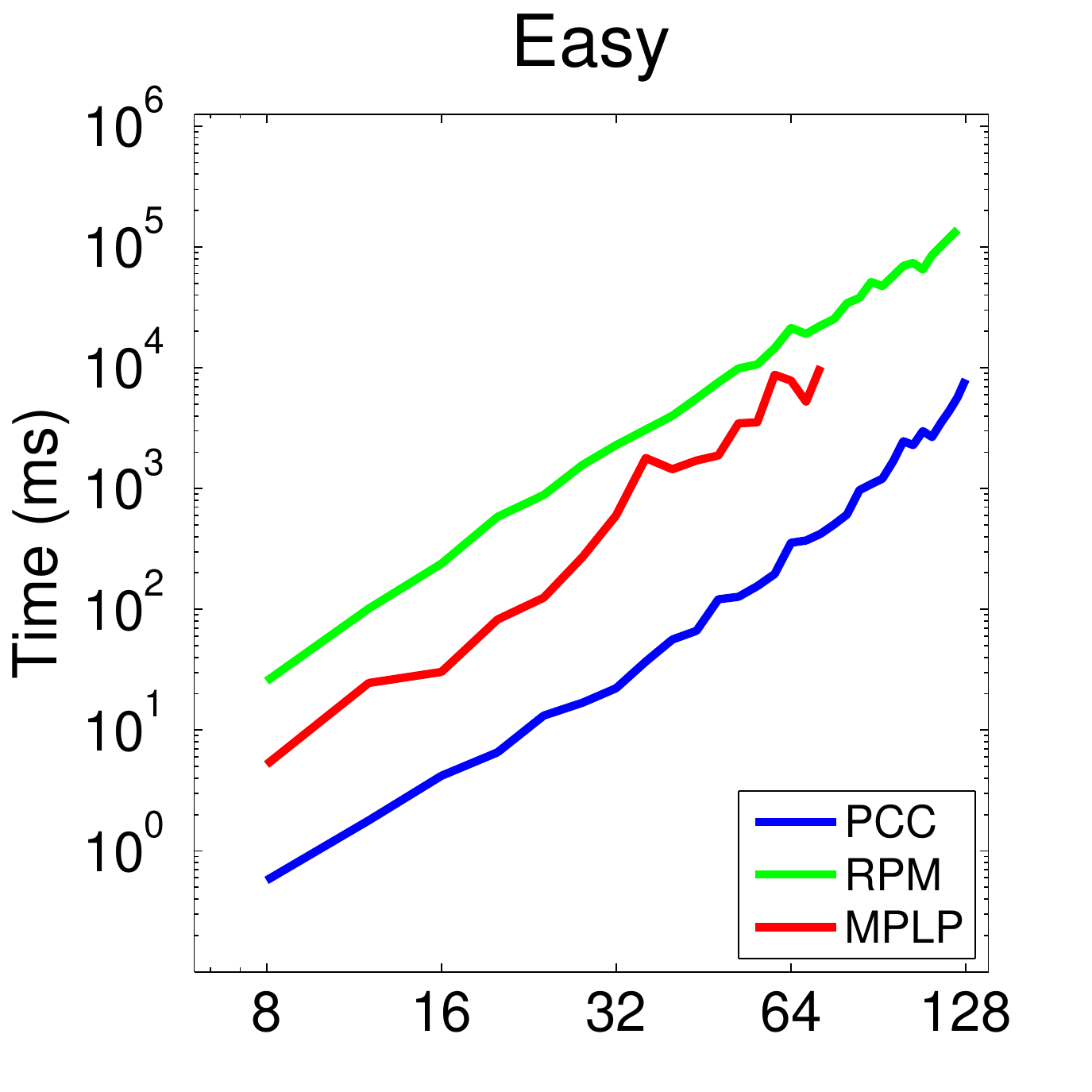}&
\hspace{-1em}\includegraphics[width=2.2in]{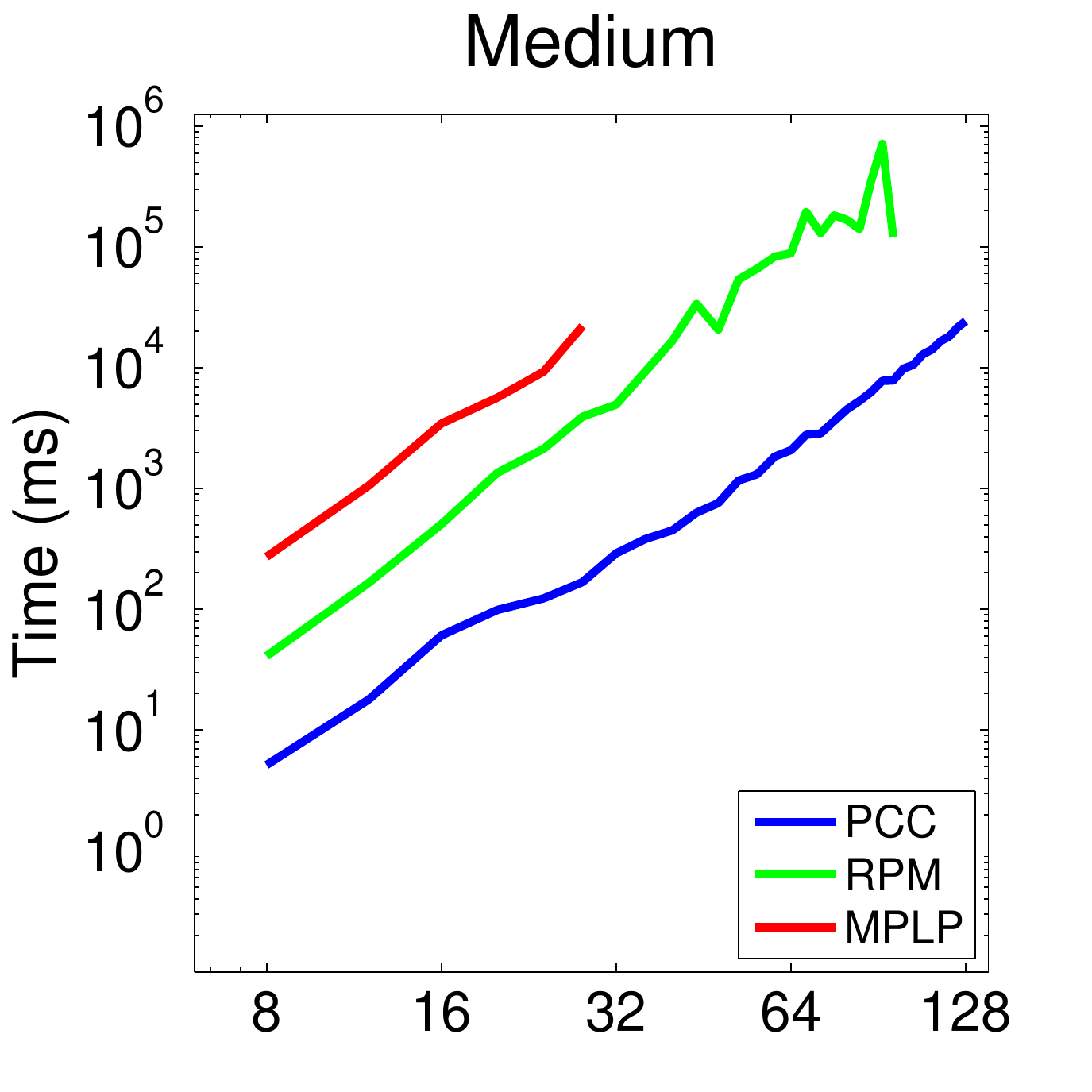}&
\hspace{-1em}\includegraphics[width=2.2in]{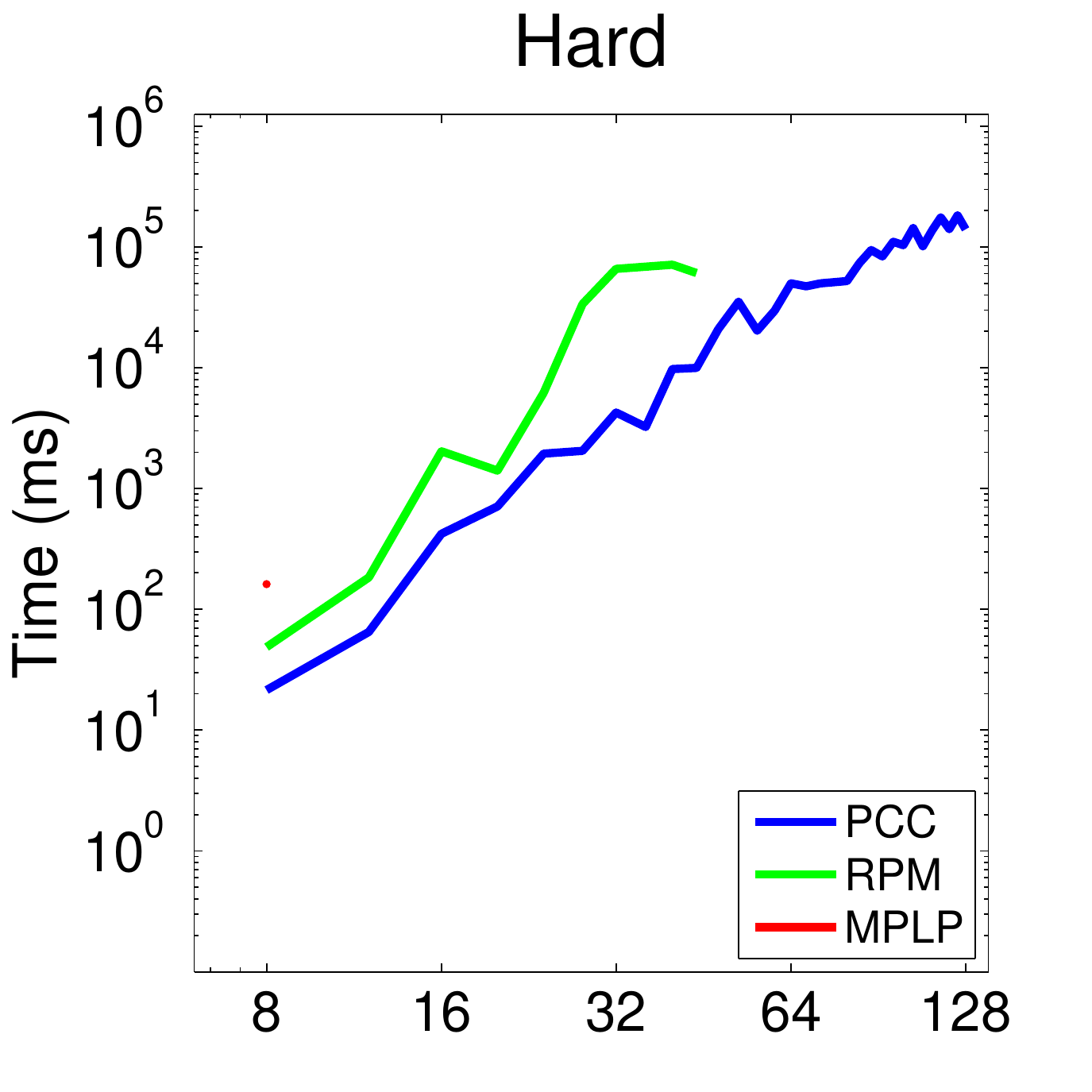}\\
\hspace{-1em}\includegraphics[width=2.07in]{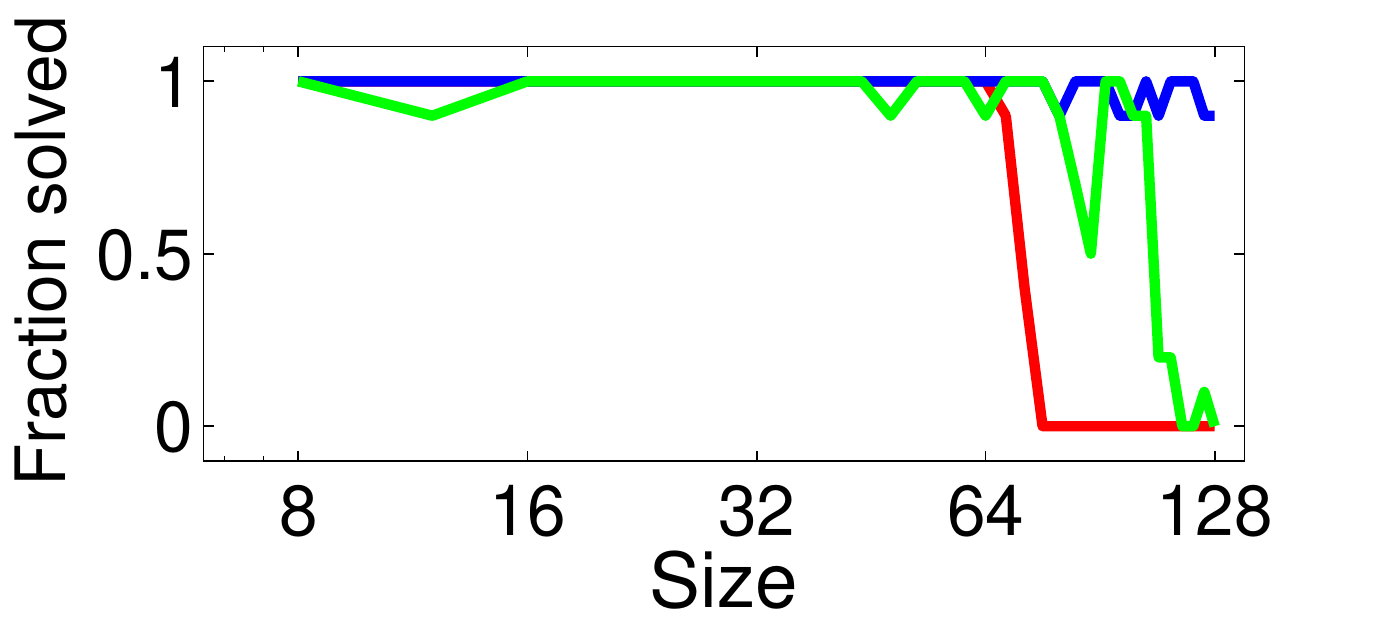}&
\hspace{-1em}\includegraphics[width=2.07in]{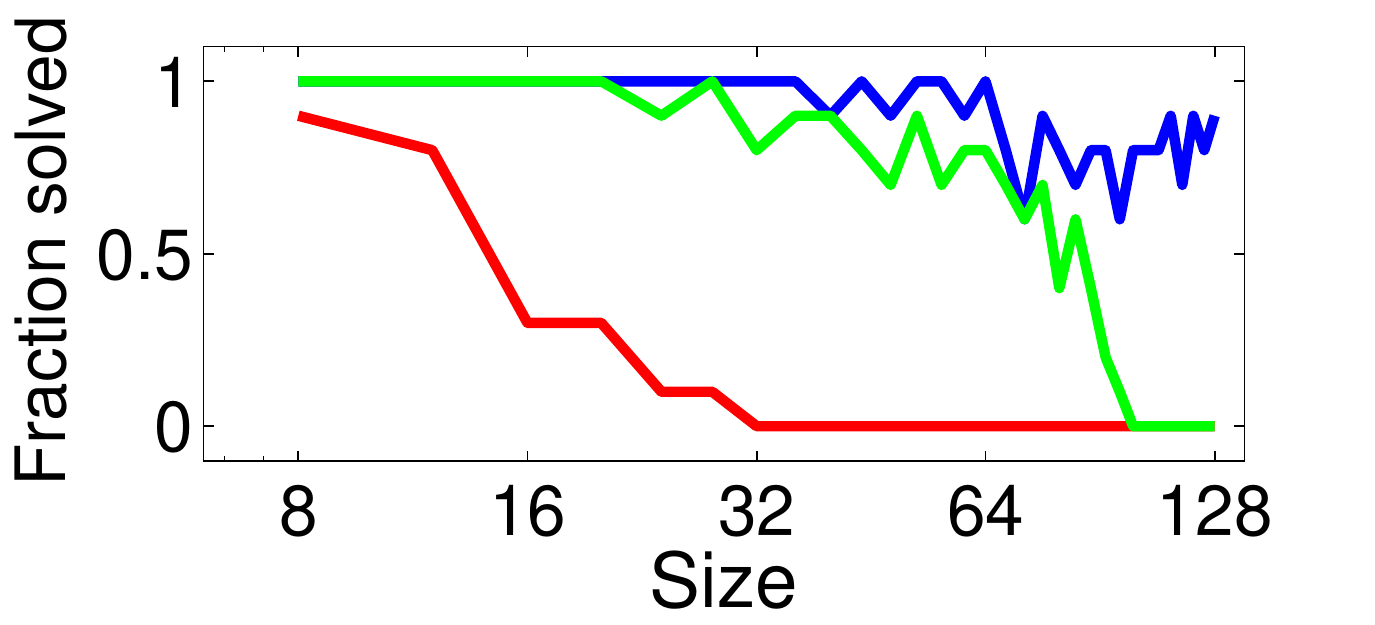}&
\hspace{-1em}\includegraphics[width=2.07in]{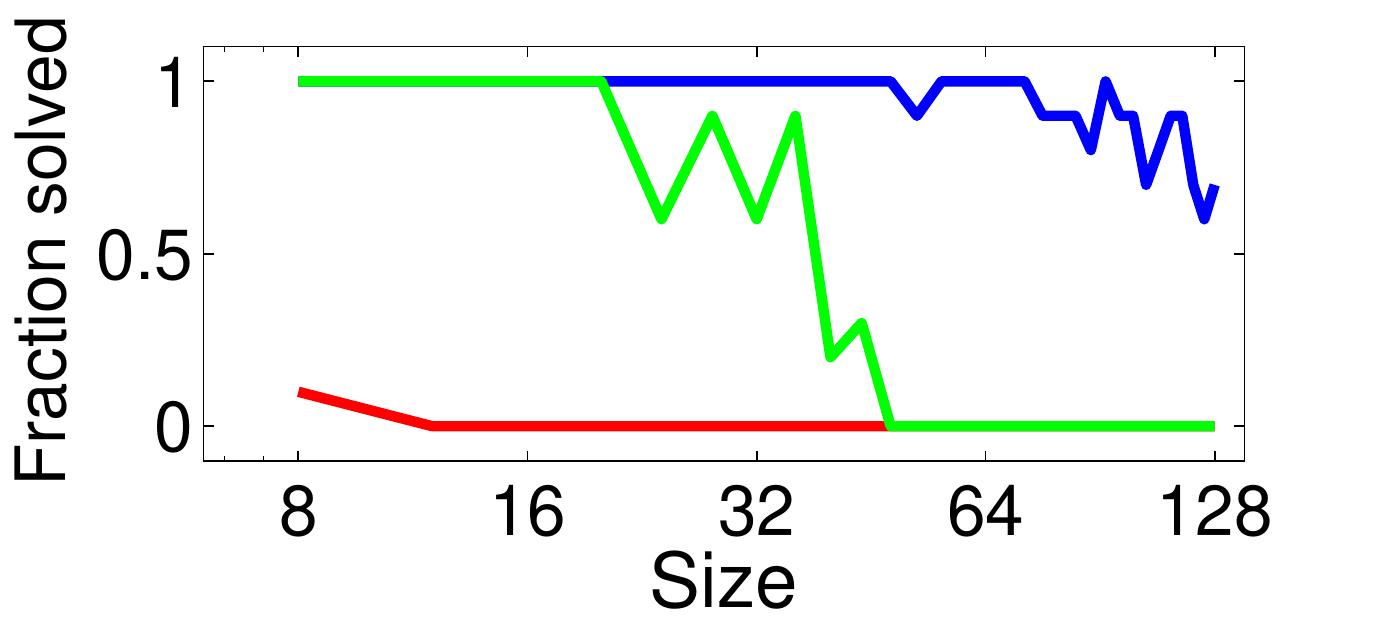}\\
\end{tabular}
\caption{Convergence times as a function of problem size for randomly generated
Ising grid problems.  We compare PCC (blue) to RPM (green) and MPLP (red) for
easy, medium and hard problems. We record times for upper- and lower- bounds to
converge averaged over 10 problem instances. We only include in the average
convergence time those problem instances for which an algorithm was able to
find the MAP configuration (a duality gap of less than 1).  The second row of
plots shows in each case the fraction of problems for which this happened.}
\label{fig:results2}
\end{figure*}

Figure~\ref{fig:results2} shows timing results as a function of problem size
for all three algorithms.  Since each method may converge (return a provably
optimal solution) on some problems but not others, we report two quantities:
the geometric mean of the time over all problems for which the method converged
(upper row), and the fraction of problems that the method successfully solved
(lower row).  As can be seen, PCC is significantly faster than the other two
methods across both problem difficulty and size, and successfully solves a
greater percentage of the problems.

\section{Discussion}

We have described a new variational bound for performing inference in planar
binary MRFs.  Our bound subsumes those given by both the tree-reweighted (TRW)
and outer-planar decompositions of such a graph since it implicitly includes
every edge and cycle as a sub-problem.  Unlike approaches such as MPLP which
successively add cycles, we are able to get the full benefit of all cycle
constraints immediately.  As a result we achieve fast convergence in practice.

The PCC graph bound is limited to planar binary problems. We are currently
exploring routes to remove these limitations.  For example, in general
non-planar graphs, we can triangulate the graph to get a cycle basis of triangles
and then ``glue'' those triangles together into the smallest possible planar
graph. In addition to MAP inference, it will also be interesting to see how the
PCC graph relates to variational approximations to the marginals.

\subsubsection*{Acknowledgements}
This work was supported by a grant from the UC Labs Research Program

{\small
\setlength{\bibspacing}{.33em}
\bibliographystyle{abbrvnat}
\bibliography{pc}
}

\end{document}